\documentclass[journal]{IEEEtran}

\usepackage{cite}
\usepackage{amsmath}
\usepackage{algorithmic}
\usepackage{array}
\usepackage{url}
\usepackage{graphicx}
\usepackage{amssymb}
\usepackage{bm}
\usepackage{caption}
\usepackage{microtype}

\begin{document}

\title{Defending Against Multiple and Unforeseen Adversarial Videos}

\author{Shao-Yuan~Lo,~\IEEEmembership{Student Member,~IEEE}
        and~Vishal M.~Patel,~\IEEEmembership{Senior Member,~IEEE}
\thanks{S.-Y.. Lo and V. M. Patel are with the Department
of Electrical and Computer Engineering, Johns Hopkins University, Baltimore, MD 21218, USA; e-mail: sylo@jhu.edu, vpatel36@jhu.edu}
}

%

\maketitle

\begin{abstract}
Adversarial robustness of deep neural networks has been actively investigated. However, most existing defense approaches are limited to a specific type of adversarial perturbations. Specifically, they often fail to offer resistance to multiple attack types simultaneously, i.e., they lack multi-perturbation robustness. Furthermore, compared to image recognition problems, the adversarial robustness of video recognition models is relatively unexplored. While several studies have proposed how to generate adversarial videos, only a handful of approaches about defense strategies have been published in the literature. In this paper, we propose one of the first defense strategies against multiple types of adversarial videos for video recognition. The proposed method, referred to as MultiBN, performs adversarial training on multiple adversarial video types using multiple independent batch normalization (BN) layers with a learning-based BN selection module. With a multiple BN structure, each BN brach is responsible for learning the distribution of a single perturbation type and thus provides more precise distribution estimations. This mechanism benefits dealing with multiple perturbation types. The BN selection module detects the attack type of an input video and sends it to the corresponding BN branch, making MultiBN fully automatic and allowing end-to-end training. Compared to present adversarial training approaches, the proposed MultiBN exhibits stronger multi-perturbation robustness against different and even unforeseen adversarial video types, ranging from Lp-bounded attacks and physically realizable attacks. This holds true on different datasets and target models. Moreover, we conduct an extensive analysis to study the properties of the multiple BN structure.
\end{abstract}


%
\IEEEpeerreviewmaketitle

\section{Introduction}
%

\IEEEPARstart{R}{ecent} advances in deep learning have led deep neural networks (DNNs) to perform outstandingly well in many computer vision problems \cite{chen2017deeplab,He_2017_ICCV,he2016deep}, including tasks such as video classification \cite{carreira2017quo,lrcn2014,hara3dcnns}. However, researchers have shown that DNNs are easily misled when presented by adversarial examples \cite{goodfellow2015explaining,Szegedy2014Intriguing}.  The adversarial examples are intentionally constructed or collected by humans to fool DNNs into making wrong predictions \cite{Goodfellow2017attacking,hendrycks2021nae}.  Most current works construct the adversarial examples by adding intentionally worst-case perturbations to input data \cite{Brown2017adversarial,goodfellow2015explaining,madry2018towards,Szegedy2014Intriguing,zajac2019adversarial}.  Various approaches have also been proposed in the literature to defend against adversarial attacks \cite{jia2019comdefend,kurakin2017adversarial,madry2018towards,raff2019barrage,samangouei2018defensegan,xie2018mitigating,Xie_2019_CVPR,addepalli2020towards}. Among them, adversarial training \cite{kurakin2017adversarial,madry2018towards,Xie_2019_CVPR} is shown to provide stronger robustness especially to the more challenging white-box and adaptive attacks \cite{obfuscated}.  Therefore, adversarial training has been used as the foundation for more advanced defense techniques. However, present adversarial training approaches usually lead to performance degradation on clean data \cite{tsipras2018robustness,zhang2019theoretically}. Xie et al. \cite{xie2020adversarial} indicated that this problem is due to the distribution mismatch between clean and adversarial examples. In order to deal with this issue, they leveraged an auxiliary batch normalization (BN) layer \cite{ioffe2015batch} to disentangle the two distributions. In addition, most existing adversarial training techniques are tailored to one specific perturbation type, e.g., a certain Lp-norm perturbation \cite{madry2018towards,raghunathan2018certified,wong2018provable} or physically realizable attacks \cite{Wu2020Defending}.  A model trained on a specific attack can improve its robustness to that particular attack but often fails to defend when presented with a sample that is perturbed by a different type of attacks \cite{sharma2018attacking}. Although there have been several attempts aim to resist multiple attack types, they usually do not consider physically realizable attacks, unforeseen attacks, or how well they perform on clean images \cite{tramer2019adv,maini2020adversarial,lin2020dual,laidlaw2021perceptual}. In a real-world application, the input data could be clean (i.e., unattacked), adversarial, or even attacked with a novel attack that the network has never seen before.

On the other hand, most recent research in this area has focused on static images. Generating adversarial examples and defense methods for videos is relatively less explored. Although a few recent works have extended adversarial attacks to videos \cite{jiang2019black,li2019stealthy,wei2019transferable,wei2019sparse,zajac2019adversarial}, we are aware of few studies so far which delve into detecting or defending against adversarial videos \cite{jia2019identifying,Xiao2019AdvITAF}. AdvIT \cite{Xiao2019AdvITAF} is one of the first adversarial frame detectors based on temporal consistency for videos. However, their approach only detects whether a video has been attacked or not. It does not provide a defense mechanism against the attacked videos. Jia et al. \cite{jia2019identifying} leveraged denoising and frame reconstruction for defense.  However, it is not clear how well their defense method works on white-box attacks as it was not reported in \cite{jia2019identifying}.

\begin{figure*}[!t]
	\centering
	\includegraphics[width=0.9\textwidth]{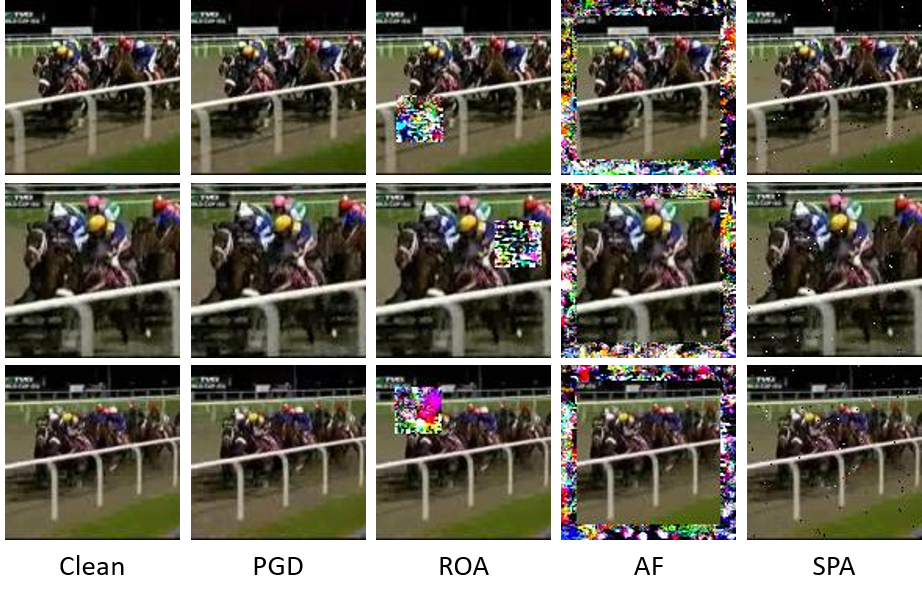}
	\caption{Illustration of the four types of adversarial videos we consider. Three video frames from the UCF-101 dataset \cite{Soomro2012ucf} are displayed here.}
	\label{fig:attacks}
\end{figure*}

In this paper, we propose MultiBN, which is one of the first defense methods for defending against adversarial videos and considering the accuracy on clean samples as well as the robustness to multiple and unforeseen perturbations. Specifically, we consider four of the most significant types of attacks: projective gradient descent (PGD) \cite{madry2018towards}, rectangular occlusion attack (ROA) \cite{Wu2020Defending}, adversarial framing (AF) \cite{zajac2019adversarial} and the proposed salt-and-pepper attack (SPA). Fig.~\ref{fig:attacks} gives an illustration of these attacks on video frames. PGD and ROA are originally designed to attack images. We extend these to videos by perturbing each frame and unveil that video recognition models are also vulnerable to these attacks. SPA is a new video attack we design, which looks like salt-and-pepper noise. PGD and SPA belong to the Lp-bounded attack group, while ROA and AF belong to the physically realizable attack group. We select one from each group as the known attack type (PGD and ROA) and leave the others as the unforeseen attack type (AF and SPA), where only the known attacks are used for adversarial training. MultiBN aims to defend against all of these attack types while retaining the performance on clean samples simultaneously.

We first demonstrate that training a model on a specific attack type can gain robustness to that attack and somewhat to another attack in the same group, but typically cannot defend against the attacks in another group. Training models on multiple attack types together (\textit{multi-perturbation training}) improves \textit{multi-perturbation robustness}, yet accuracy on clean samples is sacrificed.  This is mainly due to the distribution mismatch among clean and different types of adversarial examples. We assume that the attacks in the same group have a relatively similar distribution. Therefore, inspired by \cite{xie2020adversarial,Xie2020Intriguing}, the proposed MultiBN employs multiple BN branches in a single network: for the clean, Lp-norm and physically realizable attack examples, individually. Each BN branches is responsible for learning the distribution of a specific type of examples, which can offer more accurate distribution estimations for these types. Because BN is a lightweight component included in common DNNs, using multiple BN branches causes only minor parameter increases and computational overhead. MultiBN also contains a BN selection module, which detects the attack type of an input video and sends it to the corresponding BN branch, so the entire MultiBN is fully automatic and allows end-to-end training. Compared to existing adversarial training and multi-perturbation training approaches, MultiBN achieves stronger adversarial robustness against multiple, more diverse, and even unforeseen perturbations, while retaining higher accuracy on clean samples. Moreover, MultiBN demonstrates effectiveness in the image domain as well. An extensive analysis showing the properties of the multiple BN structure is also presented. As one of the first studies of multi-perturbation robustness for videos, this work provides baseline results that broadly cover multiple attack types, datasets and target models, for this problem. We hope that these baselines will be useful to other researchers and the adversarial robustness community.

Our main contributions are summarized as follows:
\begin{itemize}
	\item We propose a novel adversarial defense method, MultiBN, based on a multiple BN structure and a BN selection module. To the best of our knowledge, this is the first defense against multiple and unforeseen adversarial videos.
	\item The proposed MultiBN achieves both stronger multi-perturbation robustness and better clean sample performance than existing multi-perturbation training approaches. This holds true on different datasets and target models.
	\item We provide an extensive analysis to study the properties of the multiple BN structure under various conditions.
	\item We provide comprehensive baseline results for multi-perturbation robustness in the video domain. These baselines broadly cover multiple attack types, threat models, datasets, and target networks.
\end{itemize}

\section{Related Work}

\subsection{Adversarial Training}
Adversarial training is currently considered the most effective defense approach against adversarial perturbations, particularly for the white-box attacks. Goodfellow et al. \cite{goodfellow2015explaining} first proposed this strategy. They trained DNNs with both clean and adversarial images to improve adversarial robustness. Madry et al. \cite{madry2018towards} viewed adversarial training from a min-max optimization perspective, training models with solely adversarial images. It has held great promise and has been widely used as a benchmark. Zhang et al. \cite{zhang2019theoretically} introduced TRADES, which uses an alternative loss to perform adversarial training and attained better trade-off between robustness and performance. Xie et al. \cite{Xie_2019_CVPR} developed a feature denoising block, which increases the network capability of handling adversarial training. Xie et al. \cite{Xie2020Intriguing,xie2020adversarial} demonstrated that proper normalization management is important for enhancing robustness and even performance. Although our work is inspired by \cite{xie2020adversarial}, \cite{xie2020adversarial} aims to leverage the adversarial training technique to improve the image recognition performance on clean data. It does not consider the model's adversarial robustness, and its model is not applicable to the multi-perturbation robustness problem.

Several studies focus on multi-perturbation robustness. Tram{\`e}r et al. \cite{tramer2019adv} investigated adversarial robustness to multiple perturbations, including Lp-bounded attacks and rotation-translation attacks. They provided AVG and MAX adversarial training schemes. Maini et al. \cite{maini2020adversarial} incorporated multi-perturbation models into a single attack by Multi Steepest Descent (MSD). MSD is robust to different Lp-bounded attacks. Nevertheless, these studies do not take potential unforeseen attack types and clean images into consideration. Laidlaw et al. \cite{laidlaw2021perceptual} adversarially trained a target model by Neural Perceptual Threat Model (NPTM), showing good resistance to Lp-bounded attacks and spatial attacks. However, its robustness cannot generalize to physically realizable attacks. Lin et al. \cite{lin2020dual} aimed to defend against Lp and non-Lp attacks, but they require a pre-constructed On-Manifold dataset, which is too expensive for practical uses. Our MultiBN manages normalization with low costs to enhance the robustness to multiple, more diverse, and even unforeseen perturbations, while retaining higher accuracy on clean images simultaneously.

\subsection{Adversarial Videos}
Most existing literature on adversarial attacks and defense are based on static images. There are only a few works that address attacks and defense techniques for videos. Wei et al. \cite{wei2019sparse} are the first to explore adversarial examples in videos. They found that perturbations propagate through video frames in the CNN+RNN based video classifier \cite{lrcn2014}, and thus proposed a temporally sparse attack. Li et al. \cite{li2019stealthy} generated video attacks by a generative model. Zajac et al. \cite{zajac2019adversarial} developed an attack that keeps frames unchanged and just attaches an adversarial frames on the border of each video frame. Jiang et al. \cite{jiang2019black} introduced V-BAD for black-box video attacks. 

Few studies for detecting or defending against video attacks are presented. Xiao et al. \cite{Xiao2019AdvITAF} proposed AdvIT based on temporal consistency to detect adversarial frames within a video. However, their approach only detects whether a video has been attacked or not. It does not provide a defense mechanism against the attacked videos. Jia et al. \cite{jia2019identifying} presented a similar detector, along with a temporal defense and a spatial defense. The temporal defense reconstructs perturbed frames with adjacent clean frames. The spatial defense uses and denoises the reconstructed frames to mitigate the effect of adversarial perturbations. However, their approach is only evaluated on the black-box attack setting. It is not clear how well their defense method works on white-box attacks \cite{obfuscated} as it was not reported in \cite{jia2019identifying}.

\subsection{Physically Realizable Attacks}
Physical attack is a class of adversarial attacks that can be implemented in the physical space. Physically realizable attack refers to the digital representation of physical attacks. Such attacks fool DNNs by modifying physical objects being photographed. Sharif et al. \cite{Sharif16AdvML} generated printable perturbations inside eyeglass frames to attack face recognition systems. Brown et al. \cite{Brown2017adversarial} created an adversarial patch that can be put next to a real-world object, making that object be misclassified. Thys et al. \cite{thysvanranst2019} further extended the adversarial patch to fooling human detectors. Wu et al. \cite{Wu2020Defending} proposed DOA, a defense against physically realizable attacks. DOA performs adversarial training with rectangular occlusion attack (ROA), which places an adversarial rectangular sticker on an image, improving physical robustness. However, it fails to resist Lp-bounded attacks. Our MultiBN is robust to Lp-bounded attacks and physically realizable attacks simultaneously.

\section{Preliminaries}

\subsection{Multiple Adversarial Video Types}
For our investigation, we construct four types of video attacks: $L_\infty$-norm PGD \cite{madry2018towards}, ROA \cite{Wu2020Defending}, AF \cite{zajac2019adversarial}, and the new SPA attack (see Fig.~\ref{fig:attacks}). Among them, $L_\infty$-norm PGD and SPA ($L_0$-norm) belong to the Lp-bounded attacks; ROA and AF belong to the physically realizable attacks. In our experiments, we set PGD and ROA as the known attack types available for adversarial training, while AF and SPA are used as unforeseen attack types used only during inference. We aim to defend against multiple adversarial video types, including Lp-bounded and physically realizable attacks as well as known and unforeseen attacks. All of these attacks are set to untargeted since the untargeted attack is considered more difficult to resist than the targeted attack.

\subsubsection{Projective Gradient Descent}
PGD attack is defined as:
\begin{align}
x^{t+1} = \Pi_{x+\mathbb{S}}\left(x^t+\alpha \cdot sign(\bigtriangledown_x L(x,y;\theta))\right)
\end{align}
where $x$ is a data sample, $y$ is the ground-truth label, $\theta$ is model parameters, $L$ is the training loss, and $\mathbb{S}$ denotes the set of allowed perturbations. The perturbation size $\epsilon$ is described as $\parallel{x^{t_{max}}-x}\parallel_p\,\le\,\epsilon$, where $t_{max}$ denotes the maximum interation number. PGD is a powerful multi-step variant of the Fast Gradient Sign Method (FGSM) \cite{goodfellow2015explaining}. It has become one of the most important benchmarks in the current adversarial example research \cite{Wu2020Defending,Xie_2019_CVPR,Xie2020Intriguing}. We extend the $L_\infty$-norm PGD from images to videos by taking the gradient descent with respect to an entire input video.

\subsubsection{Rectangular Occlusion Attack}
ROA attack introduces $L_\infty$-norm PGD inside a fixed-size and fixed location rectangle on an image. The size is pre-defined and the location is searched with respect to the highest loss. We extend ROA to videos, in which each frame is perturbed by a rectangle. To save computations, we skip the location search step. Instead, we randomly assign the rectangle location for each video frame and then apply PGD on it.

\subsubsection{Adversarial Framing}
AF attack adds adversarial framings on the border of each video frame while most of the frame pixels are not modified. Specifically, it first fixes the framing location then applies PGD inside it. It is originally designed as a universal attack. To save computations and generate stronger adversaries, we perform a non-universal version, and it can be defined as follows:
\begin{align}
x^{t+1} = \Pi_{x+\mathbb{S}}\left(x^t+ m \cdot \alpha \cdot sign(\bigtriangledown_x L(x,y;\theta))\right)
\end{align}
where $m \in \{0, 1\}$ is the AF mask. Let $p$ be a pixel index of $\mathbf{m}$. If $p$ is on the border of $m$ within a framing width $s_{AF}$, $m_p = 1$; otherwise, $m_p = 0$.

\subsubsection{Salt-and-pepper Attack}
Inspired by the one-pixel attack \cite{Su2019one}, we design a new video attack. For computation saving, instead of using differential evolution, we randomly select a pre-defined number of pixels on each video frame, then apply PGD on those pixels. We consider it as a kind of $L_0$-norm attack because the number of adversarial pixels is bounded. We name this new attack \textit{salt-and-pepper attack (SPA)}, as the perturbations look like salt-and-pepper noise.

\subsection{Adversarial Training and Multi-perturbation Training}
The proposed MultiBN is based on adversarial training and multi-perturbation training. We briefly review adversarial training and state-of-the-art multi-perturbation training schemes to describe the preliminary formulation of our method.

To begin with, we recall the objective function for training a DNN model:
\begin{align}
\theta^* = \mathop{\arg\, \min}\limits_{\theta}\mathbb{E}_{(x,y)\sim\mathbb{D}}\left[L(x,y;\theta)\right],
\end{align}
where $x$ is a clean training sample with ground-truth label $y$ in the training set $\mathbb{D}$, $\theta$ is model parameters, and $L$ denotes the training loss.
Madry's adversarial training \cite{madry2018towards} applies the min-max optimization and trains models exclusively on adversarial examples:
\begin{align}
\theta^* = \mathop{\arg\, \min}\limits_{\theta}\mathbb{E}_{(x,y)\sim\mathbb{D}}\left[\mathop{\max}\limits_{\delta\in\mathbb{S}}L(x+
\delta,y;\theta)\right],
\end{align}
where $\delta$ denotes an adversarial perturbation in the perturbation set $\mathbb{S}$. AdaProp \cite{xie2020adversarial} aims to improve the performance on clean samples and trains the model with a mixture of clean data and adversarial examples as follows: \cite{goodfellow2015explaining,kurakin2017adversarial}:
\begin{align}
\theta^* = \mathop{\arg\, \min}\limits_{\theta}\mathbb{E}_{(x,y)\sim\mathbb{D}}\left[L(x,y;\theta)+\mathop{\max}\limits_{\delta\in\mathbb{S}}L(x+\delta,y;\theta)\right].
\end{align}
Note that AdaProp is not designed for multi-perturbation robustness. Althoug and its model does not apply to the multi-perturbation robustness problem
TRADES \cite{zhang2019theoretically} uses an alternative objective function for adversarial training:
\begin{align}
\theta^* = \mathop{\arg\, \min}\limits_{\theta}\mathbb{E}_{(x,y)\sim\mathbb{D}}\left[L(x,y;\theta)+\mathop{\max}\limits_{\delta\in\mathbb{S}}L(x+\delta,f(x);\theta)\right],
\end{align}
where $f(x)$ is the output vector of the target model with a Softmax operator. In other words, TRADES replaces $y$ with $f(x)$ to compute the cross-entropy loss of adversarial examples.

Regarding multi-perturbation robustness, Tram{\`e}r et al. \cite{tramer2019adv} introduced two adversarial training strategies: AVG strategy and MAX strategy. AVG trains on all types of adversarial examples simultaneously and optimizes these adversarial losses together as follows:
\begin{align} \label{avg}
\theta^* = \mathop{\arg\, \min}\limits_{\theta}\mathbb{E}_{(x,y)\sim\mathbb{D}}\left[\sum_{i=1}^{N}\mathop{\max}\limits_{\delta_i\in\mathbb{S}_i}L(x+\delta_i,y;\theta)\right],
\end{align}
where $N$ is the number of perturbation types. MAX considers the worst-case attack. It trains on the strongest adversarial example that obtains the maximum loss among all types of attacks: 
\begin{align} \label{max}
\theta^* = \mathop{\arg\, \min}\limits_{\theta}\mathbb{E}_{(x,y)\sim\mathbb{D}}\left[L(x+\delta_k,y;\theta)\right],
\end{align}
where
\begin{align}
\delta_k=\arg\mathop{\max}\limits_{i\in[1,N]}\left[\mathop{\max}\limits_{\delta_i\in\mathbb{S}_i}L(x+\delta_i,y;\theta)\right],
\end{align}
which denotes the strongest type of attacks. MSD \cite{maini2020adversarial} maximizing the worst-case loss over all the considered perturbations at each projected steepest descent step to construct a single perturbation. This can be described as the follows:
\begin{align} \label{msd}
\theta^* = \mathop{\arg\, \min}\limits_{\theta}\mathbb{E}_{(x,y)\sim\mathbb{D}}\left[L(x+\delta_{MSD},y;\theta)\right],
\end{align}
where $\delta_{MSD}$ is the constructed single perturbation.

\section{Proposed Method}
\begin{figure*}
	\centering
	\includegraphics[width=1\textwidth]{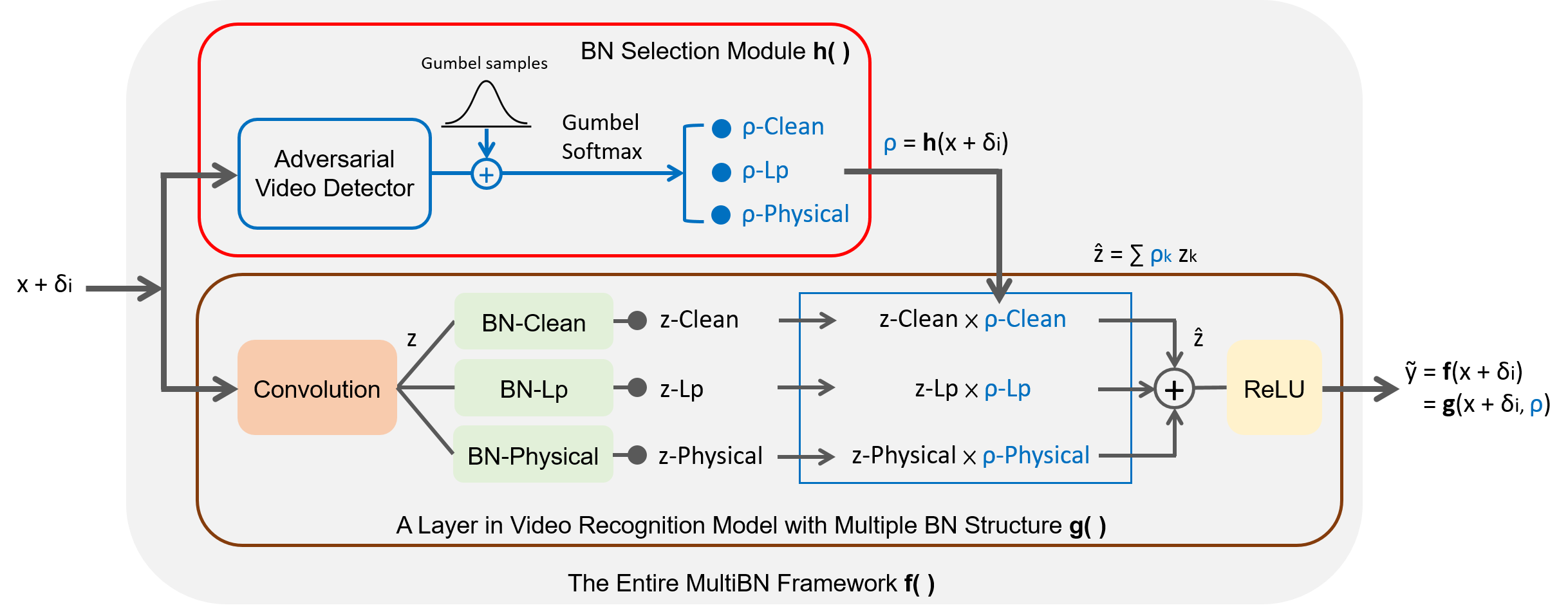}
	\caption{Overview of MultiBN, the proposed adversarial defense framework. Every batch normalization layer of the video recognition model is replaced by a multiple BN structure, where this figure illustrates only one layer for simplicity. $x + \delta_i$: an input of a specific type adversarial example, $z_k$: the $k$-th BN branch's output features, $\rho_k$: a ratio factor to weight the $k$-th BN branch's output features, $\tilde{y}$: prediction.}
	\label{fig:framework}
\end{figure*}

In real-world applications, the input data could be clean, adversarial, or even attacked with a novel attack that the network has never seen before. Hence, it is important to design a defense solution that can resist multiple known and unforeseen perturbations while retaining the performance on clean samples. The proposed method, MultiBN, is based on a multiple BN structure and a BN selection module. Fig.~\ref{fig:framework} gives an overview of MultiBN.

\subsection{Multiple Batch Normalization Structure}
\label{sec:method_a}
Adversarial training on a single perturbation type is generally weakly robust to the other types of attacks. On the other hand, most state-of-the-art DNNs contain BNs \cite{ioffe2015batch} in their architecture to normalize input features, which improves performance \cite{hara3dcnns,he2016deep}. However, owing to the different distributions among multiple perturbation types, BNs suffer from the distribution mismatch when multi-perturbation training is conducted, and thus fails to gain promising multi-perturbation robustness. To address this problem, we deploy multiple BN branches into each BN layer of the target model and keep the rest of the parts unchanged, i.e., still a single network \cite{xie2020adversarial,Xie2020Intriguing}. Clean data and each perturbation type used for training are assigned an individual BN branch. Since BN is a lightweight component, multiple BN branches cause only minor parameter increases and computational overhead.

Ideally, each BN branch is responsible for estimating the assigned a single or a family distribution(s), and thus can properly disentangle multiple distributions. Consider adversarial training as a min-max optimization problem \cite{madry2018towards}, for forward and backward passes, we can manually make each perturbation type attack the target model through its assigned BN branch at the inner maximization step. For the outer minimization step, we send clean inputs or the generated adversarial examples to their corresponding BN branch as well. The ideal objective function can be defined as follows:
\begin{align} \label{eqmine}
\begin{split}
\theta^* = & \mathop{\arg\, \min}\limits_{\theta}\mathop{\mathbb{E}}\limits_{(x,y)\sim\mathbb{D}} \\ & \left[L(x,y;\theta^c,\theta^b_0)+\sum_{i=1}^{N}\mathop{\max}\limits_{\delta_i\in\mathbb{S}_i}L(x+\delta_i,y;\theta^c,\theta^b_i)\right],
\end{split}
\end{align}
where $\theta^c$ is convolution parameters, $\theta_i^b$ is the BN parameters of the $i$-th data type, and $\theta=\theta^c+\sum_{i=0}^{N}\theta_i^b$ denotes all the model parameters.

In practical scenarios, DNNs should provide robustness against unforeseen attacks. An exhaustive investigation is too expensive; instead, we can summarize different attack types into several groups based on their distributions, then build a BN branch for each group. In our case, we deploy three BN branches for clean data, Lp-bounded attacks and physically realizable attacks, respectively (see Fig.~\ref{fig:framework}). Typically, adversarial training on a strong attack has better robustness \cite{madry2018towards}, where PGD and ROA are good representatives of Lp-bounded attacks and physically realizable attacks, respectively. Therefore, we train a target model on clean, PGD, and ROA examples using the 3-BN structure with Eq.~\eqref{eqmine}, where $N$ = 2.

\subsection{Batch Normalization Selection Module}
\label{sec:method_b}
At inference time, we cannot control the data flow, so the input data have to pass through the corresponding BN branch automatically. To this end, we propose a BN selection module based on an adversarial video detector and a Gumbel-Softmax operator \cite{jang2017categorical,maddison2017concrete} (see Fig.~\ref{fig:framework}). The adversarial video detector is achieved by a $(N+1)$-class video classification model, where $N$ is the number of attack types used for training. It is trained to not only identify whether an input video is clean or adversarial but also recognize the attack types. In our case, the detector is trained with $N=2$ on clean, PGD and ROA examples to recognize clean data, Lp-bounded attacks and physically realizable attacks. Hence, unforeseen perturbations would also be classified into the most similar attack group.

Intuitively, we can build a switch module to send the input to the proper BN branch according to its detection result. However, the $argmax$ operation, which applies to the adversarial video detector's logits for BN branch selection, is not differentiable. This makes end-to-end training infeasible. To address this issue, we leverage the Gumbel-Softmax trick to allow the gradients to backpropagate through a discrete sampling process \cite{xu2020learning}. Specifically, we approximate $argmax$ by the differentiable and continuous Gumbel-Softmax function, which is defined as follows:
\begin{align}
\rho_i = \frac{exp((\log\,\pi_i + G_i) \, / \, \tau)}{\sum_{j=1}^{K} exp((\log\,\pi_j + G_j) \, / \, \tau)},
\end{align}
where $\pi_1,...,\pi_K$ are the adversarial video detector's logits, $G_1,...,G_K$ are i.i.d. Gumbel samples, $\tau$ is the Softmax temperature, and $K=3$ in our case. Next, $\rho_1,...,\rho_K$ are used as ratio factors to weight each BN branch's output features:
\begin{align} \label{bn_aggregate}
\hat{z} = \sum_{i=1}^{K} \rho_k \, z_k,
\end{align}
where $z_1,...,z_K$ are each BN branch's output features, and $\hat{z}$ is the weighted feature that would be the input of the next network component (see Fig.~\ref{fig:framework}). In this way, the correct BN branch's output feature $z_*$ would dominate $\hat{z}$, making $\hat{z}$ be a good approximation of $z_*$.

\subsection{The Entire Framework}
With the BN selection module, MultiBN can operate automatically during inference without manual control, and it enables end-to-end training since the entire framework is differentiable. Let $f(\,)$ be the entire MultiBN framework, $g(\,)$ be the video recognition model with the multiple BN structure, and $h(\,)$ be the BN selection module (see Fig.~\ref{fig:framework}). Consider an input of a specific type adversarial example $x + \delta_i$, the entire end-to-end pipeline can be described as follows:
\begin{align} \label{pipeline}
\begin{split}
\tilde{y} & = f(x + \delta_i; \, \theta^{c}, \theta^{b}, \theta^{det}) \\
& = g(x + \delta_i, h(x + \delta_i; \, \theta^{det}); \, \theta^{c}, \theta^{b}),
\end{split}
\end{align}
where $\tilde{y}$ is the prediction, $\theta^{c}$ is $g(\,)$'s convolution parameters, $\theta^{b} = \sum_{i=0}^{N}\theta_i^b$ ($N = 2$ here, see Subsection~\ref{sec:method_a}) is $g(\,)$'s BN parameters in all the BN branches, and $\theta^{det}$ denotes the parameters of the adversarial video detector in the BN selection module $h(\,)$. $h(\,)$ outputs $\bm{\rho} = [\rho_1,...,\rho_K]$ defined in Subsection~\ref{sec:method_b}, i.e., $\bm{\rho} = h(x + \delta_i; \, \theta^{det})$.

Then, the end-to-end training objective can be written as follows:
\begin{align} \label{comb_eq}
\begin{split}
\theta^* & = \mathop{\arg \, \min}\limits_{\theta} \mathop{\mathbb{E}}\limits_{(x,y)\sim\mathbb{D}} \Big[ L(x,y;\,\theta) + \lambda \cdot L(x,y^{det};\,\theta^{det}) \\ & + \sum_{i=1}^{N} \big( \mathop{\max}\limits_{\delta_i\in\mathbb{S}_i}L(x+\delta_i,y;\,\theta) + \lambda \cdot L(x+\delta_i,y^{det};\,\theta^{det}) \big) \Big],
\end{split}
\end{align}
where $\theta = \theta^{c} + \theta^{b} + \theta^{det}$ contains all the entire MultiBN's parameters, $y$ is task (video recognition here) ground-truth, $y^{det}$ is the ground-truth of the video types for training the adversarial video detector, $L$ is the usual cross-entropy loss, and $\lambda$ is a trade-off hyperparameter. The objectives $L(x,y;\theta)$ and $L(x,y^{det};\theta^{det})$ are trained for clean data, while $L(x+\delta_i,y;\theta)$ and $L(x+\delta_i,y^{det};\theta^{det})$ are for adversarial training. $L(x,y^{det};\theta^{det})$ and $L(x+\delta_i,y^{det};\theta^{det})$ exclusively learns the BN selection module $h(\,)$, while $L(x,y;\theta)$ and $L(x+\delta_i,y;\theta)$ learns the entire framework $f(\,)$ in an end-to-end manner.

\subsection{Defense Mechanism Against Unforeseen Attacks}
In Subsection~\ref{sec:method_a} and \ref{sec:method_b}, we mention how the proposed method addresses the presence of unforeseen attacks during inference. Here, we further elaborate on its mechanism.

We deal with unforeseen attacks via proper attack type categorization, which is achieved by the proposed multiple BN structure and BN selection module. We consider PGD, ROA, AF and SPA attacks in this paper. Suppose that we are aware of only PGD and ROA at training time. We classify PGD and ROA to the Lp-bounded attack group and the physically realizable attack group, respectively, based on the perturbation distributions. The MultiBN framework is built according to this categorization. Specifically, we deploy three BN branches for clean data, Lp-bounded attacks and physically realizable attacks, respectively (see Fig.~\ref{fig:framework}). Each BN is adversarially trained to be robust against each particular category, and the BN selection module is trained to identify the most similar category of given input data.

Suppose that AF and SPA are unforeseen to us at training time but present at test time. During inference, the BN selection module identifies the most similar category of the input AF and SPA examples (i.e., the physically realizable attack group and the Lp-bounded attack group, respectively). Accordingly, the features from their belonging BN branch would dominate the feature maps after the feature aggregation step described in Eq.~(\ref{bn_aggregate}). Since each BN is robust against a particular attack group, MultiBN can achieve high robustness against the unforeseen AF and SPA attacks, which are classified as the most similar group by the BN selection module.

In contrast, a model without the multiple BN structure cannot be uniformly robust against multiple attack groups (see Subsection~\ref{evaluation}), resulting in sub-optimal robustness against unforeseen attacks. Besides, a model without the BN selection module cannot properly aggregate the features from different BN branches. Therefore, the proposed MultiBN framework consisting of the multiple BN structure and the BN selection module can decently address unforeseen attacks.

\section{Experiments}
In this section, we first describe our experimental setup. Second, we evaluate MultiBN's manually-controlled version to validate the effectiveness of the multiple BN structure and explore the properties of this structure. Next, we test the proposed MultiBN's robustness and performance and compare it with state-of-the-art multi-perturbation training approaches. We also evaluate MultiBN's robustness against adaptive attacks, different attack budgets, and black-box attacks. Finally, we conduct further analyses on model size, sanity checks, and the experiments on images.

\renewcommand{\arraystretch}{1.2}
\setlength{\tabcolsep}{12pt}
\begin{table*}
	\begin{center}
		\caption{Results (\%) of MultiBN-manual on target model 3D ResNeXt-101 and dataset UCF-101. No Defense is trained on only clean data. AT-PGD, AT-ROA, AT-AF and AT-SPA are adversarially trained on a single specific attack type. The best results are in bold, and the best results among adversarially trained models are underlined.}
		\label{table:pre_results}
		\begin{tabular}{l | r | rrrr | rr}
			\hline \noalign{\smallskip} \noalign{\smallskip}
			Model & Clean & PGD & ROA & AF & SPA & Mean & Union \\
			\noalign{\smallskip} \hline \noalign{\smallskip}
			No Defense & \textbf{89.0} & 3.3 & 0.5 & 1.6 & 8.4 & 20.6 & 0.0 \\
			\noalign{\smallskip} \hline \noalign{\smallskip}
			AT-PGD & 78.6 & \textbf{49.0} & 5.0 & 0.6 & 67.1 & 40.1 & 0.3 \\
			AT-ROA & 82.6 & 12.5 & \textbf{69.0} & 54.0 & 17.6 & 47.1 & 7.9 \\
			AT-AF & 84.6 & 7.1 & 3.9 & \textbf{80.5} & 12.2 & 37.7 & 2.1 \\
			AT-SPA & 83.5 & 36.9 & 2.6 & 0.7 & \textbf{69.5} & 38.6 & 0.2 \\
			\noalign{\smallskip} \hline \noalign{\smallskip}
			MultiBN-manual & \underline{83.7} & \underline{46.4} & \underline{65.6} & \underline{57.0} & \underline{60.4} & \textbf{62.6} & \textbf{40.7} \\
			\noalign{\smallskip} \hline
		\end{tabular}
	\end{center}
\end{table*}

\setlength{\tabcolsep}{10pt}
\begin{table}
	\begin{center}
		\caption{Results (\%) of each BN branch on the five input types. BN-Clean, BN-Lp and BN-Physical are the clean, PGD and ROA BN branches in the multiple BN structure, respectively.}
		\label{table:branch}
		\begin{tabular}{l | rrrrr}
			\hline \noalign{\smallskip} \noalign{\smallskip}
			BN Branch & Clean & PGD & ROA & AF & SPA \\
			\noalign{\smallskip} \hline \noalign{\smallskip}
			BN-Clean & \textbf{83.7} & 21.3 & 13.5 & 5.9 & 23.8 \\
			BN-Lp & 79.0 & \textbf{46.4} & 7.7 & 1.9 & \textbf{60.4} \\
			BN-Physical & 83.0 & 23.5 & \textbf{65.6} & \textbf{57.0} & 26.6 \\
			\noalign{\smallskip} \hline
		\end{tabular}
	\end{center}
\end{table}

\setlength{\tabcolsep}{12pt}
\begin{table*}
	\begin{center}
		\caption{Results (\%) of the cases that the target BN and the inference BN are different.}
		\label{table:infattack}
		\begin{tabular}{l | rrr | rrr}
			\hline \noalign{\smallskip} \noalign{\smallskip}
			Attack & \multicolumn{2}{r}{\underline{\quad PGD \quad}} & &  \multicolumn{2}{r}{\underline{\quad ROA \quad}} &  \\
			Inference BN $\backslash$ Target BN & BN-Clean & BN-Lp & BN-Physical & BN-Clean & BN-Lp & BN-Physical \\
			\noalign{\smallskip} \hline \noalign{\smallskip}
			BN-Clean & 21.3 & 50.9 & 35.9 & 13.5 & 17.9 & 56.6 \\
			BN-Lp & \textbf{72.6} & 46.4 & \textbf{70.5} & 30.4 & 7.7 & 48.2 \\
			BN-Physical & 46.4 & \textbf{52.3} & 23.5 & \textbf{78.5} & \textbf{76.4} & \textbf{65.6} \\
			\noalign{\smallskip} \hline
			\noalign{\smallskip} \noalign{\smallskip}
			Attack & \multicolumn{2}{r}{\underline{\quad AF \quad}} & &  \multicolumn{2}{r}{\underline{\quad SPA \quad}} & \\
			Inference BN $\backslash$ Target BN & BN-Clean & BN-Lp & BN-Physical & BN-Clean & BN-Lp & BN-Physical \\
			\noalign{\smallskip} \hline \noalign{\smallskip}
			BN-Clean & 5.9 & 7.3 & 49.0 & 23.8 & 55.7 & 41.8 \\
			BN-Lp & 16.4 & 1.9 & 33.5 & \textbf{77.2} & \textbf{60.4} & \textbf{75.8} \\
			BN-Physical & \textbf{75.2} & \textbf{62.5} & \textbf{57.0} & 49.8 & 57.4 & 26.6 \\
			\noalign{\smallskip} \hline
		\end{tabular}
	\end{center}
\end{table*}

\setlength{\tabcolsep}{12pt}
\begin{table*}
	\begin{center}
		\caption{Results (\%) of MultiBN and state-of-the-art approaches on target model 3D ResNeXt-101 and dataset UCF-101. The best results are in bold.}
		\label{table:main_results}
		\begin{tabular}{l | r | rrrr | rr}
			\hline \noalign{\smallskip} \noalign{\smallskip}
			Model & Clean & PGD & ROA & AF & SPA & Mean & Union \\
			\noalign{\smallskip} \hline \noalign{\smallskip}
			No Defense & \textbf{89.0} & 3.3 & 0.5 & 1.6 & 8.4 & 20.6 & 0.0 \\
			\noalign{\smallskip} \hline \noalign{\smallskip}
			TRADE \cite{zhang2019theoretically} (ICML'19) & 82.3 & 29.0 & 5.7 & 3.3 & 42.2 & 32.5 & 1.9 \\
			AVG \cite{tramer2019adv} (NeurIPS'19) & 68.9 & 38.1 & 51.4 & 18.5 & 49.6 & 45.3 & 17.3 \\
			MAX \cite{tramer2019adv} (NeurIPS'19) & 72.8 & 32.5 & 31.0 & 5.8 & 49.4 & 38.3 & 5.5 \\
			MSD \cite{maini2020adversarial} (ICML'20) & 70.2 & 43.2 & 1.7 & 1.6 & \textbf{56.0} & 34.6 & 0.7 \\
			\noalign{\smallskip} \hline \noalign{\smallskip}
			MultiBN (ours) & 74.2 & \textbf{44.6} & \textbf{58.6} & \textbf{44.3} & 53.7 & \textbf{55.1} & \textbf{34.8} \\
			\noalign{\smallskip} \hline
		\end{tabular}
	\end{center}
\end{table*}

\setlength{\tabcolsep}{12pt}
\begin{table*}
	\begin{center}
		\caption{Results (\%) of MultiBN and state-of-the-art approaches on target model 3D Wide ResNet-50 and dataset UCF-101. The best results are in bold.}
		\label{table:wideresnet}
		\begin{tabular}{l | r | rrrr | rr}
			\hline \noalign{\smallskip} \noalign{\smallskip}
			Model & Clean & PGD & ROA & AF & SPA & Mean & Union \\
			\noalign{\smallskip} \hline \noalign{\smallskip}
			No Defense & \textbf{88.4} & 11.5 & 0.2 & 1.0 & 10.0 & 22.2 & 0.0 \\
			\noalign{\smallskip} \hline \noalign{\smallskip}
			TRADE \cite{zhang2019theoretically} (ICML'19) & 81.1 & 26.7 & 1.1 & 0.7 & 39.2 & 29.8 & 0.1 \\
			AVG \cite{tramer2019adv} (NeurIPS'19) & 74.5 & 43.1 & 55.6 & 3.5 & 57.2 & 46.8 & 3.5 \\
			MAX \cite{tramer2019adv} (NeurIPS'19) & 76.0 & 32.5 & 12.2 & 2.3 & 39.2 & 32.4 & 1.9 \\
			MSD \cite{maini2020adversarial} (ICML'20) & 71.0 & 46.3 & 2.9 & 0.9 & \textbf{61.1} & 36.4 & 0.2 \\
			\noalign{\smallskip} \hline \noalign{\smallskip}
			MultiBN (ours) & 77.4 & \textbf{46.5} & \textbf{59.9} & \textbf{48.1} & 56.7 & \textbf{57.7} & \textbf{37.8} \\
			\noalign{\smallskip} \hline
		\end{tabular}
	\end{center}
\end{table*}

\setlength{\tabcolsep}{12pt}
\begin{table*}
	\begin{center}
		\caption{Results (\%) of MultiBN and state-of-the-art approaches on target model 3D ResNeXt-101 and dataset HMDB-51. The best results are in bold.}
		\label{table:hmdb51}
		\begin{tabular}{l | r | rrrr | rr}
			\hline \noalign{\smallskip} \noalign{\smallskip}
			Model & Clean & PGD & ROA & AF & SPA & Mean & Union \\
			\noalign{\smallskip} \hline \noalign{\smallskip}
			No Defense & \textbf{65.1} & 0.0 & 0.0 & 0.0 & 0.3 & 13.1 & 0.0 \\
			\noalign{\smallskip} \hline \noalign{\smallskip}
			TRADE \cite{zhang2019theoretically} (ICML'19) & 54.8 & 6.8 & 0.3 & 0.0 & 20.5 & 16.5 & 0.0 \\
			AVG \cite{tramer2019adv} (NeurIPS'19) & 39.0 & 14.3 & 17.1 & 2.8 & 26.2 & 19.9 & 1.4 \\
			MAX \cite{tramer2019adv} (NeurIPS'19) & 48.6 & 13.9 & 16.0 & 0.1 & 30.3 & 21.8 & 0.0 \\
			MSD \cite{maini2020adversarial} (ICML'20) & 41.4 & 18.2 & 0.1 & 0.0 & \textbf{31.2} & 18.2 & 0.0 \\
			\noalign{\smallskip} \hline \noalign{\smallskip}
			MultiBN (ours) & 51.1 & \textbf{22.0} & \textbf{23.7} & \textbf{7.8} & 29.9 & \textbf{26.9} & \textbf{5.0} \\
			\noalign{\smallskip} \hline
		\end{tabular}
	\end{center}
\end{table*}

\subsection{Experimental Setup}

\subsubsection{Datasets}
We use UCF-101 \cite{Soomro2012ucf} and HMDB-51 \cite{kuehne2011hmdb} for evaluation, which are widely used video datasets in action recognition. UCF-101 consists of 13,320 videos from 101 action classes, and HMDB-51 has 6,766 videos from 51 action classes. Following \cite{wei2019sparse}, we resize their frame dimensinos to $112\times112$ and uniformly sample each video into 40 frames. UCF-101 is the default dataset if not otherwise specified.

\subsubsection{Attack Setting}
We consider $L_\infty$-norm PGD \cite{madry2018towards}, ROA \cite{Wu2020Defending}, AF \cite{zajac2019adversarial}, and the proposed SPA attack. For PGD, we set the perturbation size $\epsilon$ to $4/255$; for ROA, we set the rectangle size $s_{ROA}$ to $30\times30$ and $\epsilon$ to $255/255$; for AF, we set the framing width $s_{AF}$ to $10$ and $\epsilon$ to $255/255$; for SPA, we set the number of adversarial pixels on each frame $s_{SPA}$ to $100$ and $\epsilon$ to $255/255$. The number of attack iterations $t_{max}$ is set to $5$ for all the four attacks. To test the proposed method by strong attacks, all of these attacks are untargeted attacks and in the white-box setting (i.e., the attacker has full knowledge of the target model, including the multiple BN structure and the BN selection module).

\subsubsection{Implementation Details}
We choose 3D ResNext-101 and 3D Wide ResNet-50 \cite{hara3dcnns} as our target models, as they are two of the most top-performing 3D CNNs for video recognition, where 3D ResNext-101 is the default target model if not otherwise specified. For our adversarial video detector, we choose the lightweight 3D ResNet-18. We use the pre-trained weights from \cite{hara3dcnns} and conduct adversarial training upon the pre-trained models. We set MultiBN's Softmax temperature $\tau=1$ and the trade-off hyperparameter $\lambda=0.1$. All the models are trained by a SGD optimizer with initial learning rate $5e^{-4}$, momentum $0.9$ and weight decay $1e^{-5}$, where the learning rate is decreased by a factor of $10$ in the middle of the training process.

We apply the mean accuracy and the union accuracy as the metrics to evaluate the multi-perturbation robustness. The union accuracy requires that the target models need to correctly classify an input sample under all the considered input types.

\subsection{Multiple BN Structure} \label{evaluation}
We first manually select the correct BN branches to investigate the effectiveness of the multiple BN structure. We call this variant MultiBN-manual. Then, we compare MultiBN-manual with vanilla adversarial training \cite{madry2018towards} that trains on a single attack type. Table~\ref{table:pre_results} shows that models trained on a specific attack always have the best robustness to that attack. AT-PGD and AT-ROA also yield high robustness to another attack in their own group, showing better generalization. However, all of them almost fail to defend against the attacks from other groups. 

MultiBN-manual uniformly achieves the second-highest accuracy across all the five input types, and most of these accuracies are close to the best one. Although MultiBN-manual is not the best from the perspective of any specific input type, it sustains a much better balance when multiple input types are considered. This shows the effectiveness of the multiple BN structure in multi-perturbation robustness. As can be seen, MultiBN-manual's mean accuracy and union accuracy are significantly higher than adversarial training on a single attack.

\subsection{Analysis of Differnt BN Branches}
In the previous subsection, the attacker generates perturbations through the BN branch corresponding to its type. During inference, the input is also sent to that BN branch accordingly. In this subsection, we investigate the cases that perturbations are produced on a BN branch which is different from the group of the used attack type, and the input is sent to that BN branch during inference.

In Table~\ref{table:branch}, we can see that each BN branch performs the best on the input type which they are trained on. Moreover, for the unforeseen attacks, BN-Lp is the most robust to SPA, and BN-Physical is the most robust to AF. This result is consistent to our assumption that feeding an unforeseen adversarial example to the BN branch of the same or the most similar group can enjoy the best benefit. Our grouping follows the observation that PGD and SPA have similar distributions (Lp-bounded attacks), and ROA and AF have similar distributions (physically realizable attacks).

\subsection{Analysis of Target BN and Inference BN}
We further delve into the cases that an adversarial example is made inference on a BN branch (inference BN) different from the BN branch that is used to generate the adversarial example (target BN). In other words, we consider the cases that the target BN and the inference BN are different.

The results in Table~\ref{table:infattack} are mostly consistent with that in Table~\ref{table:branch}, in which BN-Lp has the strongest robustness to Lp-bounded attacks, and BN-Physical has the strongest robustness to physically realizable attacks. PGD attack is an exception: When the target BN is BN-Lp, inference BN-Lp performs the worst. 

In addition, we observe that for any specific inference BN, it is more robust to the adversarial examples generated on another BN branch, i.e., target BN and inference BN are different. In such a case, the attack is not a rigorous white-box attack, so we treat it as a kind of gray-box attack, in which the attacker does not know which BN branch would the adversarial example pass through during inference. This unveils that the attacks cannot perfectly transfer to other BN branches though the rest of the model parameters are shared in the same network. Such results show that the multiple BN structure can make secure against this gray-box setting.

\subsection{Robustness of the Entire MultiBN Framework}
In this subsection, we evaluate the entire MultiBN and compare it with state-of-the-art adversarial training and multi-perturbation training approaches, including TRADES \cite{zhang2019theoretically}, AVG \cite{tramer2019adv}, MAX \cite{tramer2019adv} and MSD \cite{maini2020adversarial}. For TRADES, we apply the AVG strategy to it for multi-perturbation training. Because we take clean data accuracy into consideration, we adjust AVG, MAX and MSD by involving clean data in training. That is, we add the clean data loss term $L(x,y;\theta)$ into the expectation of objective functions Eq.~(\ref{avg}), Eq.~(\ref{max}) and Eq.~(\ref{msd}).

Table~\ref{table:main_results} reports the results on target model 3D ResNext-101 and dataset UCF-101, Table~\ref{table:wideresnet} reports the results on target model 3D Wide ResNet-50 and dataset UCF-101, Table~\ref{table:hmdb51} reports the results on target model 3D ResNext-101 and dataset HMDB-51. As expected, No Defense still achieves the best performance on clean data, showing that adversarial training degrades clean data performance. TRADES has the best clean data performance among the adversarial training approaches, but it lacks multi-perturbation robustness. AVG improves multi-perturbation robustness to a large extent, yet its clean data performance is very low. MAX is less robust than AVG in our case. MSD has the best and the second-best robustness against SPA and PGD, respectively, but it is vulnerable to physically realizable attacks. The proposed MultiBN achieves the second-best clean data performance among the defenses, the second-best robustness against SPA, and the best robustness against the rest of the attack types. MultiBN consistently outperforms the competitors in terms of mean accuracy and union accuracy by a wide margin, showing great multi-perturbation robustness. This holds true on different datasets and target models.

\begin{figure*}
	\centering
	\captionsetup{justification=centering}
	\includegraphics[width=1\textwidth]{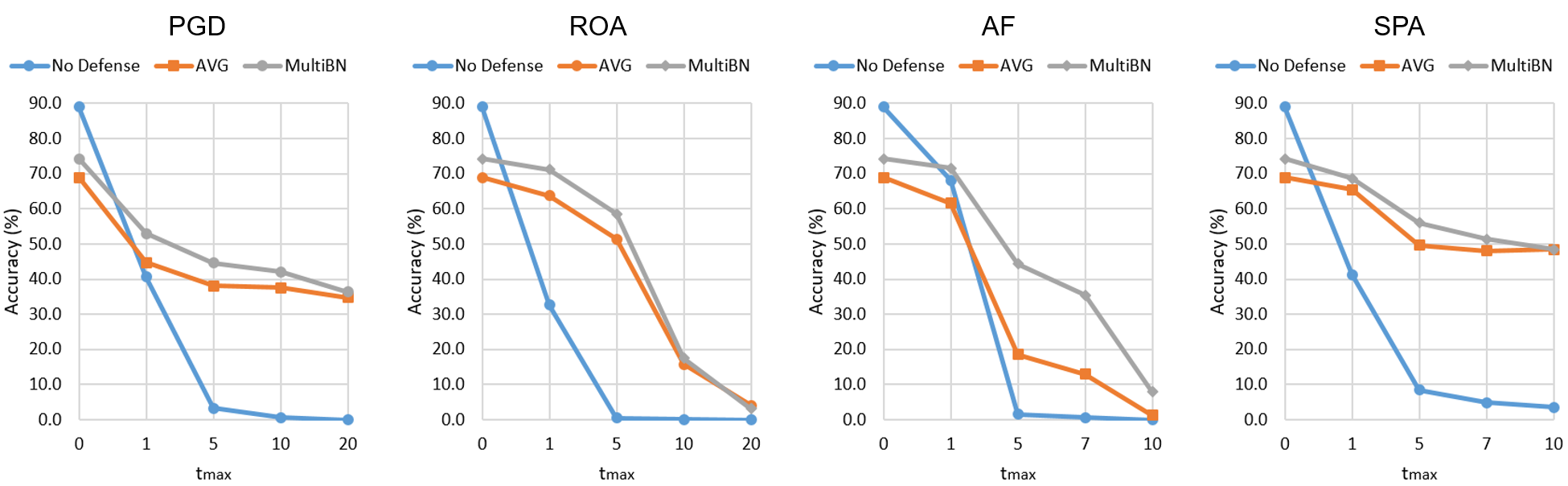}
	\caption{Results (\%) under the four attack types with varied numbers of attack iterations $t_{max}$.}
	\label{fig:tmax}
\end{figure*}

\begin{figure*}
	\centering
	\captionsetup{justification=centering}
	\includegraphics[width=1\textwidth]{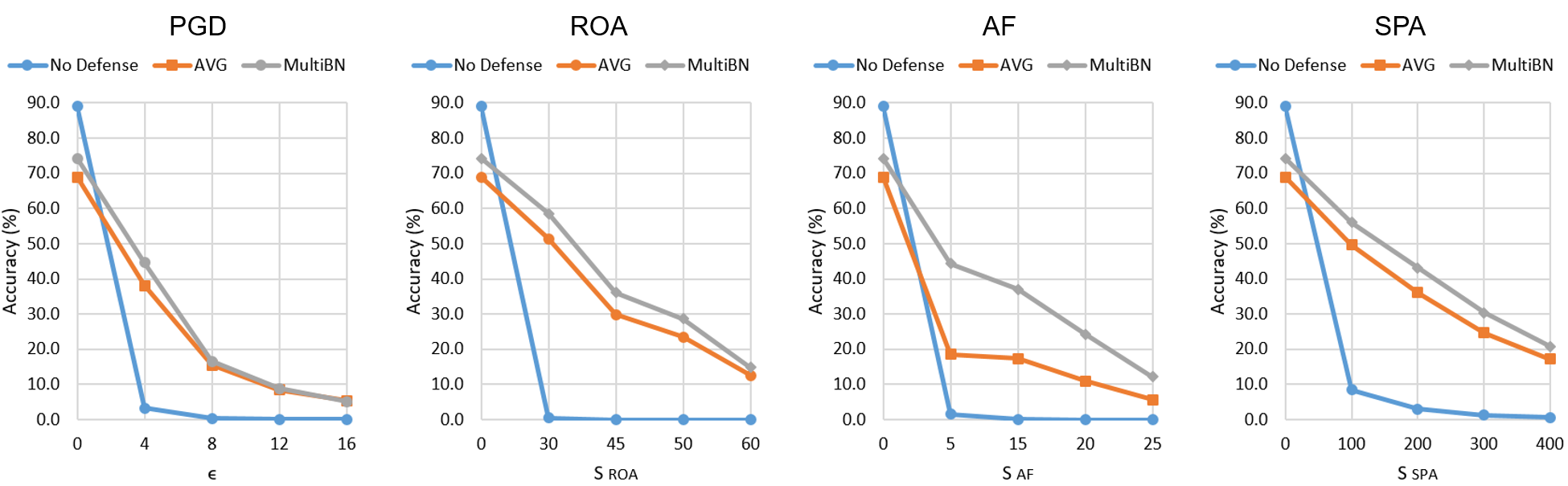}
	\caption{Results (\%) under the four attack types with varied perturbation bounds.}
	\label{fig:epsilon}
\end{figure*}

\begin{figure}
	\centering
	\includegraphics[width=0.4\textwidth]{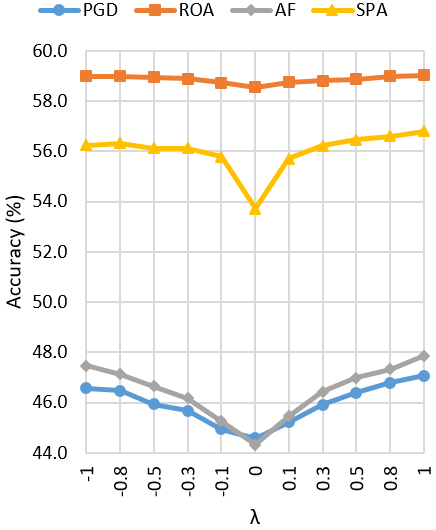}
	\caption{Results (\%) of MultiBN under the adaptive attacks with varied $\lambda$ of the four attack types.}
	\label{fig:adaptive}
\end{figure}

\subsection{Attack Budgets}
To further evaluate the effectiveness of MultiBN, we test its scalability to different attack budgets. We vary the attack budgets by two aspects: The number of attack iterations $t_{max}$ and the perturbation bounds of different attack types, i.e., PGD's $\epsilon$, ROA's $s_{ROA}$, AF's $s_{AF}$ and SPA's $s_{SPA}$. The results are presented in Fig.~\ref{fig:tmax} and Fig.~\ref{fig:epsilon}. The No Defense baseline and the strongest competitor AVG are compared.

We can see that MultiBN consistently achieves better robustness against different attack types with various attack iterations and perturbation bounds. This shows that MultiBN's multi-perturbation robustness is scalable to various attack budgets.

\subsection{Robustness Against Adaptive Attacks}
To thoroughly evaluate MultiBN, we construct an adaptive attack \cite{tramer2020adaptive}, which jointly attacks the target model part and the BN selection module part. The intuition is to generate adversarial examples which can also fool the BN selection module to let it select the incorrect BN branch, and thus become easier to fool the target model. This adaptive attack is formulated as follows:
\begin{align}
\delta=\arg \mathop{\max}\limits_{\delta\in\mathbb{S}} \big[ L(x+\delta,y;\theta) + \lambda \cdot L(x+\delta,y^{det};\theta^{det}) \big].
\end{align}

As presented in Fig.~\ref{fig:adaptive}, the canonical attack has the greatest attacking strength. The accuracies under all the four attack types monotonously increase as $|\lambda|$ increases. This shows that the considered adaptive attack fails to break MultiBN.

\subsection{Robustness Against Black-box Attacks}
In addition to the white-box robustness we discussed, we also evaluate the proposed method's robustness against black-box attacks \cite{papernot2017practical}. Table~\ref{table:black_box} reports the results on UCF-101. Here, we consider a naturally trained (i.e., train with only clean data) 3D Wide ResNet-50 as a substitute model to generate black-box adversarial examples, and test on the target model, 3D ResNeXt-101. As we can see, the proposed MultiBN uniformly achieves excellent robustness against multiple attack types in the black-box setting. In particular, MultiBN's robust accuracies are very close to its clean accuracy (74.2\%), showing that the black-box attacks hardly fool it. Its union accuracy attains 63.5\%, which significantly outperforms all the competitors.

\subsection{Model Size Analysis}
\begin{figure}
	\centering
	\includegraphics[width=0.45\textwidth]{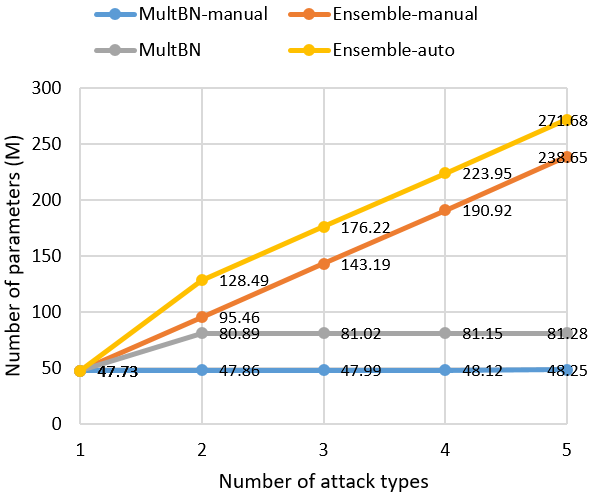}
	\caption{Model size analysis result. ``-manual" refers to the model without the BN selection module.}
	\label{fig:model_size}
\end{figure}

Apart from performance and robustness, model size is another critical factor when we evaluate a model. This regards the feasibility of a model for real-world applications. Our MultiBN significantly improves multi-perturbation robustness with only a minor increase in the number of parameters. To present the compactness of the MultiBN architecture, we construct a naive model ensemble approach as a baseline for comparison. The model ensemble approach trains an individual model for each particular attack type, and uses our BN selection module to select the corresponding model for the input video during inference. Fig.~\ref{fig:model_size} compares the number of parameters of MultiBN and the model ensemble. The model ensemble's number of parameters linearly increases along with the number of attack types since its number of individual models equals the number of attack types. In contrast, MultiBN only deploys distinct BN parameters for each particular attack type and shares all the rest of the parameters across all the attack types. Hence, the increase of model size is minimal, especially compared to the backbone network's size. This demonstrates that the proposed method obtains excellent effectiveness and model compactness simultaneously.

\subsection{Sanity Checks to Evaluation}
To verify whether the proposed MultiBN's robustness is not due to obfuscated gradients, we report our results on the basic sanity checks introduced by Athalye et al. \cite{obfuscated}.
\begin{itemize}
	\item Fig.~\ref{fig:tmax} shows that iterative attacks are stronger than one-step attacks.
	\item Table~\ref{table:main_results} and Table~\ref{table:black_box} show that white-box attacks are stronger than black-box attacks.
	\item Unbounded attacks reach 100.0\% attack success rate (accuracy drops to 0.0\%).
	\item Fig.~\ref{fig:epsilon} shows that increasing distortion bound increases attack success (decreases accuracy). 
\end{itemize}
These results confirm that our MultiBN's robustness is indeed not due to obfuscated gradients, which further demonstrates its reliability.

\setlength{\tabcolsep}{15pt}
\begin{table*}
	\begin{center}
		\caption{Results (\%) of MultiBN and state-of-the-art approaches under black-box attacks on UCF-101. The substitute model is the naturally trained 3D Wide ResNet-50, and the target model is 3D ResNeXt-101. The best results are shown in bold, and the second-best results are underlined.}
		\label{table:black_box}
		\begin{tabular}{l | r | rrrr | r}
			\hline \noalign{\smallskip} \noalign{\smallskip}
			Model & Clean & PGD & ROA & AF & SPA & Union \\
			\noalign{\smallskip} \hline \noalign{\smallskip}
			TRADE \cite{zhang2019theoretically} (ICML'19) & \textbf{82.3} & \textbf{81.0} & 60.8 & \underline{65.0} & \textbf{78.0} & 49.3 \\
			AVG \cite{tramer2019adv} (NeurIPS'19) & 68.9 & 68.4 & 68.0 & 62.0 & 68.4 & 56.2 \\
			MAX \cite{tramer2019adv} (NeurIPS'19) & 72.8 & 72.4 & \underline{71.4} & 63.5 & \underline{71.9} & \underline{57.9} \\
			MSD \cite{maini2020adversarial} (ICML'20) & 70.2 & 69.8 & 40.1 & 52.2 & 69.1 & 31.3 \\
			\noalign{\smallskip} \hline \noalign{\smallskip}
			MultiBN (ours) & \underline{74.2} & \underline{73.6} & \textbf{74.0} & \textbf{72.4} & 71.5 & \textbf{63.5} \\
			\noalign{\smallskip} \hline
		\end{tabular}
	\end{center}
\end{table*}

\setlength{\tabcolsep}{12pt}
\begin{table*}
	\begin{center}
		\caption{Results (\%) of MultiBN and state-of-the-art approaches on target model ResNet-18 and dataset CIFAR-10. The best results are shown in bold, and the second-best results are underlined.}
		\label{table:image_results}
		\begin{tabular}{l | r | rrrr | rr}
			\hline \noalign{\smallskip} \noalign{\smallskip}
			Model & Clean & PGD & ROA & AF & SPA & Mean & Union \\
			\noalign{\smallskip} \hline \noalign{\smallskip}
			No Defense & \textbf{94.3} & 0.0 & 4.7 & 0.1 & 16.3 & 23.1 & 0.0 \\
			\noalign{\smallskip} \hline \noalign{\smallskip}
			TRADE \cite{zhang2019theoretically} (ICML'19) & 71.4 & 14.7 & 34.7 & 30.4 & 52.8 & 40.8 & 10.1 \\
			AVG \cite{tramer2019adv} (NeurIPS'19) & 86.4 & 47.2 & 53.6 & 60.5 & \underline{67.8} & 63.1 & 28.1 \\
			MAX \cite{tramer2019adv} (NeurIPS'19) & 87.7 & 46.3 & 60.0 & 54.6 & \textbf{73.6} & \underline{64.4} & \underline{33.7} \\
			MSD \cite{maini2020adversarial} (ICML'20) & 93.0 & \textbf{52.7} & 6.7 & 7.1 & 59.6 & 43.8 & 2.2 \\
			\noalign{\smallskip} \hline \noalign{\smallskip}
			MultiBN (ours) & \underline{94.2} & \underline{49.7} & \textbf{74.9} & \textbf{66.7} & 60.9 & \textbf{69.3} & \textbf{36.9} \\
			\noalign{\smallskip} \hline
		\end{tabular}
	\end{center}
\end{table*}

\subsection{Results on Images}
The proposed method is effective in the image domain as well. For the experiment on images, we use CIFAR-10 \cite{krizhevsky2009learning} as the dataset and ResNet-18 \cite{he2016deep} as the target model. The architecture of the adversarial video detector is also ResNet-18. Regarding attack setting, we set the perturbation size $\epsilon$ to $8/255$ for PGD, the rectangle size $s_{ROA}$ to $12\times12$ for ROA, the framing width $s_{AF}$ to $3$ for AF, and the number of adversarial pixels on each image $s_{SPA}$ to $30$ for SPA. All the attacks are untargeted attacks and in the white-box setting.

Table~\ref{table:image_results} reports the evaluation results. Compared to the state-of-the-art approaches, MultiBN achieves the best accuracy under the ROA and AF attacks and the second-best accuracy under clean images and the PGD attack. Similar to the results on videos, MultiBN is far superior to all the competitors in terms of mean accuracy and union accuracy. This demonstrates that MultiBN can be a preferred solution for multi-perturbation robustness in both the image and the video domains.

\section{Conclusions}
In this paper, we proposed MultiBN, a new adversarial defense method aiming at multi-perturbation robustness. This is one of the first defenses against multiple and unforeseen adversarial videos. MultiBN uses a multiple BN structure to solve the distribution mismatch problem during multi-perturbation training. A BN selection module makes the entire framework automatic at inference time and differentiable for end-to-end training. Compared to existing adversarial training approaches, MultiBN achieves stronger multi-perturbation robustness against different and even unforeseen Lp-bounded attacks and physically realizable attacks. This holds true on different datasets and target models. Furthermore, we conduct an extensive analysis to explore the properties of the multiple BN structure under various conditions. In our future work, we will consider video-specific properties, such as temporal information, for adversarial attacks and defenses in videos.

\section*{Acknowledgment}
This work was supported by the DARPA GARD Program HR001119S0026-GARD-FP-052.

{\small
\bibliographystyle{IEEEtran}
\bibliography{mycite}
}

%
\newpage

\begin{IEEEbiography}[{\includegraphics[width=1in,height=1.25in,clip,keepaspectratio]{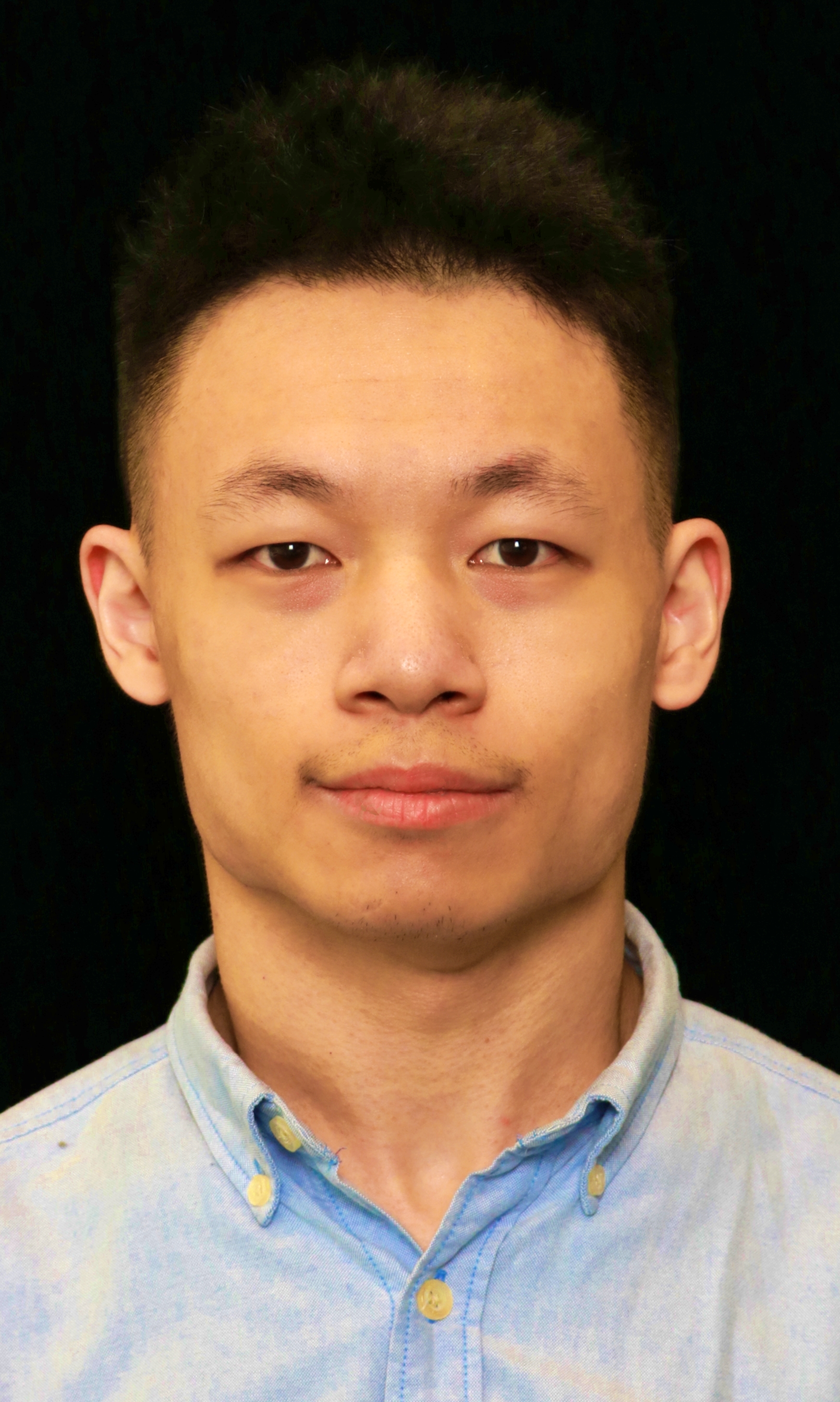}}]{Shao-Yuan Lo}
(Student Member, IEEE) is a Ph.D. student in the Department of Electrical and Computer Engineering at Johns Hopkins University. He received his B.S. and M.S. degrees from National Chiao Tung University, Taiwan, in 2017 and 2019, respectively. His research interests include adversarial machine learning, domain adaptation and semantic segmentation. He received the Best Paper Award at ACM Multimedia Asia 2019 and the 2019 IPPR Best Master Thesis Award.
\end{IEEEbiography}

\begin{IEEEbiography}[{\includegraphics[width=1in,height=1.25in,clip,keepaspectratio]{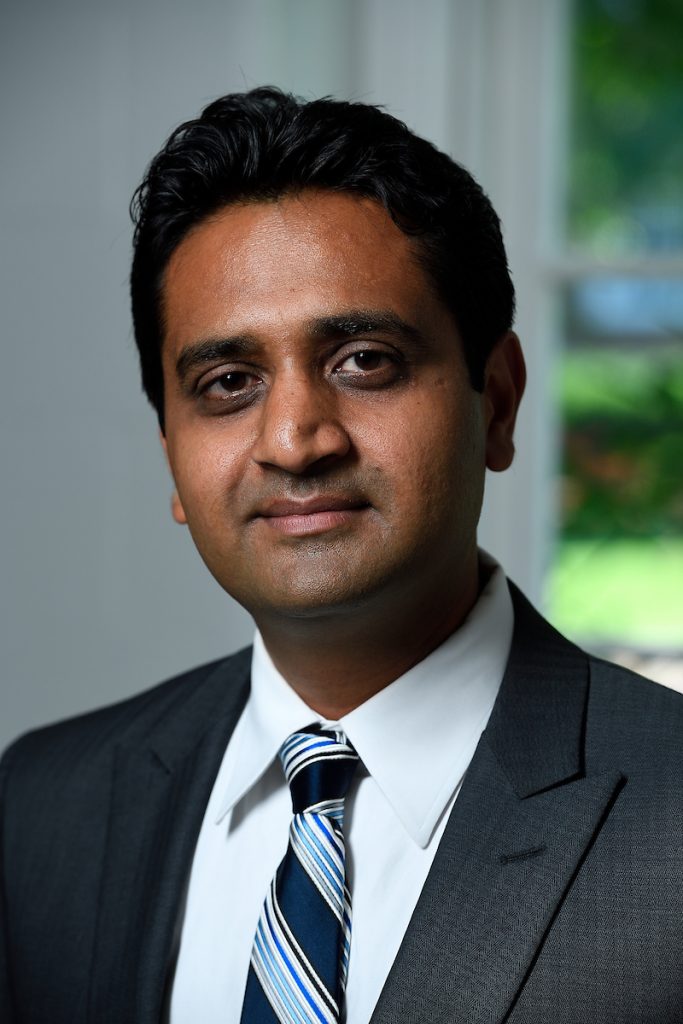}}]{Vishal M. Patel}
(Senior Member, IEEE) is an Associate Professor in the Department of Electrical and Computer Engineering (ECE) at Johns Hopkins University. Prior to joining Hopkins, he was an A. Walter Tyson Assistant Professor in the Department of ECE at Rutgers University and a member of the research faculty at the University of Maryland Institute for Advanced Computer Studies (UMIACS). He completed his Ph.D. in Electrical Engineering from the University of Maryland, College Park, MD, in 2010. He has received a number of awards including the 2021 NSF CAREER Award, the 2021 IAPR Young Biometrics Investigator Award (YBIA), the 2016 ONR Young Investigator Award, the 2016 Jimmy Lin Award for Invention, A. Walter Tyson Assistant Professorship Award, Best Paper Awards at IEEE AVSS 2017 \& 2019, IEEE BTAS 2015, IAPR ICB 2018, IEEE ICIP 2021, and two Best Student Paper Awards at IAPR ICPR 2018. He is an Associate Editor of the IEEE Transactions on Pattern Analysis and Machine Intelligence, Pattern Recognition Journal, and serves on the Machine Learning for Signal Processing (MLSP) Committee of the IEEE Signal Processing Society. He serves as the vice president of conferences for the IEEE Biometrics Council.
\end{IEEEbiography}

\vfill


\end{document}